\def\year{2019}\relax
\documentclass[letterpaper]{article} 
\pdfoutput=1
\usepackage{aaai19}  
\usepackage{times}  
\usepackage{helvet}  
\usepackage{courier}  
\usepackage{url}  
\usepackage{graphicx}  
\usepackage{amsmath}
\usepackage{tabularx,tabulary}
\usepackage{algorithm2e}
\usepackage{algpseudocode}
\usepackage{amsmath,amssymb} 

\usepackage[pagebackref=true,breaklinks=true,colorlinks,bookmarks=false]{hyperref}
\newcommand{\etal}{\textit{et al}.}

\usepackage{amsmath}
\usepackage{bm}

\DeclareRobustCommand{\uvec}[1]{{%
  \ifcsname uvec#1\endcsname
     \csname uvec#1\endcsname
   \else
    \bm{\hat{\mathbf{#1}}}%
   \fi
}}

\usepackage{array}
\newcolumntype{L}[1]{>{\raggedright\let\newline\\\arraybackslash\hspace{0pt}}m{#1}}
\newcolumntype{C}[1]{>{\centering\let\newline\\\arraybackslash\hspace{0pt}}m{#1}}
\newcolumntype{R}[1]{>{\raggedleft\let\newline\\\arraybackslash\hspace{0pt}}m{#1}}

\usepackage{xspace}
\makeatletter
\DeclareRobustCommand\onedot{\futurelet\@let@token\@onedot}
\def\@onedot{\ifx\@let@token.\else.\null\fi\xspace}

\def\etal{\emph{et al}\onedot}
\makeatother

\begin{document}
%
\title{BiHMP-GAN: Bidirectional 3D Human Motion Prediction GAN}
\author{Jogendra Nath Kundu\thanks{equal contribution - listed alphabetically by first names} \qquad Maharshi Gor\footnotemark[1] \qquad R. Venkatesh Babu\\
Video Analytics Lab, Department of Computational and Data Sciences\\
Indian Institute of Science, Bangalore, India\\
{\tt\small jogendrak@iisc.ac.in, maharshigor18@gmail.com, venky@iisc.ac.in}}
\maketitle
\begin{abstract}
Human motion prediction model has applications in various fields of computer vision. Without taking into account the inherent stochasticity in the prediction of future pose dynamics, such methods often converges to a deterministic undesired mean of multiple probable outcomes. Devoid of this, we propose a novel probabilistic generative approach called Bidirectional Human motion prediction GAN, or \textit{BiHMP-GAN}. To be able to generate multiple probable human-pose sequences, conditioned on a given starting sequence, we introduce a random extrinsic factor $r$, drawn from a predefined prior distribution. Furthermore, to enforce a direct content loss on the predicted motion sequence and also to avoid mode-collapse, a novel bidirectional framework is incorporated by modifying the usual discriminator architecture. The discriminator is trained also to regress this extrinsic factor $r$, which is used alongside with the intrinsic factor (encoded starting pose sequence) to generate a particular pose sequence. To further regularize the training, we introduce a novel recursive prediction strategy. In spite of being in a probabilistic framework, the enhanced discriminator architecture allows predictions of an intermediate part of pose sequence to be used as a conditioning for prediction of the latter part of the sequence. The bidirectional setup also provides a new direction to evaluate the prediction quality against a given test sequence. For a fair assessment of \textit{BiHMP-GAN}, we report performance of the generated motion sequence using (i) a critic model trained to discriminate between real and fake motion sequence, and (ii) an action classifier trained on real human motion dynamics. Outcomes of both qualitative and quantitative evaluations, on the probabilistic generations of the model, demonstrate the superiority of \textit{BiHMP-GAN} over previously available methods.

\end{abstract}

\begin{figure}
\begin{center}
	\includegraphics[width=1.0\linewidth]{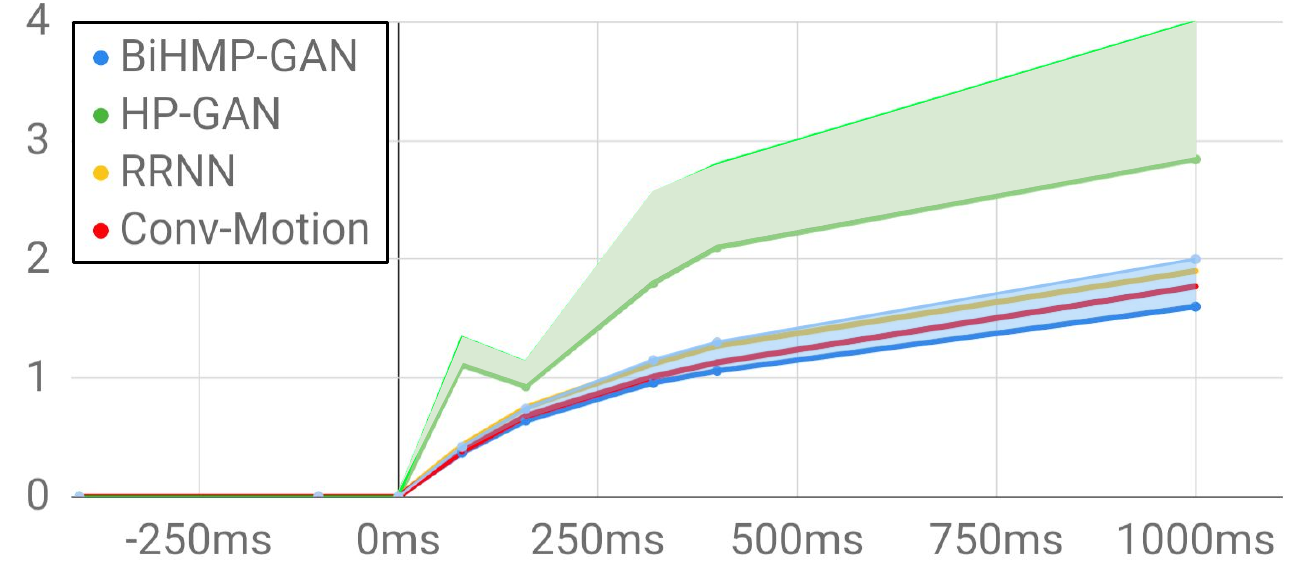}
	\caption{Mean average prediction error on Human 3.6M for different motion prediction methods. The blue and green band for \textit{BiHMP-GAN} and HP-GAN respectively, show uncertainty in the prediction of future motion as compared to other deterministic approaches.}
    \vspace{-6mm}
 	\label{fig:fig_1}    
\end{center}
\end{figure}

\section{Introduction}
\noindent Seamless interaction of robot or AI systems with urban environment dominated by human beings requires certain behaviour prediction abilities. In this work, the focus is on understanding the dynamics of human pose. For example, the ability to predict pedestrian behaviour in an urban road scene is very crucial for autonomous driving systems to prevent potential accidents. Other examples include interaction of robots with humans; such as handshaking, catching or holding objects thrown by other person etc. Moreover the artificial systems must develop the ability to understand the general trends of human pose dynamics for effective and coherent interactions~\cite{koppula2013anticipating}. Humans develop such ability by observing actions or pose dynamics of other persons over time. Creating a system which models such diverse human actions is the prime motive towards achieving an efficient human motion prediction model~\cite{mainprice2013human}.

The goal is to develop a model, which can predict plausible 3D human pose sequence from a given past dynamics of a certain time period. However, prediction of future pose sequence should not be modeled as a deterministic approach as there can be multiple plausible limb variations conditioned on the past motion dynamics. Since, the uncertainty in probable future pose increases with increase in time, a deterministic model cannot be considered reliable for long-term predictions. For example a person running may slow down to stop or keep running at a different speed. Although such variations are present in the available human motion dataset, some pose dynamics are more probable than other. Hence, the model should have the flexibility to model such stochasticity in the prediction of future pose sequence.

With the advent of deep learning for sequence-to-sequence~\cite{sutskever2014sequence} modeling, many recent works use variants of deep recurrent neural networks for human motion prediction and synthesis~\cite{ghosh2017learning,li2017auto}. According to the analysis performed by Martinez \etal~\cite{Martinez_2017_CVPR}, earlier motion prediction methods~\cite{taylor2007modeling} show a catastrophic drift in the prediction of immediate future frame conditioned on past motion sequence. They proposed to solve it by utilizing the recurrent network to predict the residue on past frame instead of directly estimating the next frame parameters. However most of the recent works in human motion prediction~\cite{li2018convolutional,butepage_2017_CVPR} do not model the inherent stochasticity in the fore-casted pose sequence. In such scenario, the model predicts a deterministic undesired mean of multiple probable pose dynamics, which often leads to suboptimal performance. We address this issue by introducing a randomly sampled vector (or an extrinsic factor) along with the latent representation of encoded past frames - the intrinsic representation. We consider the combination of these two factors as the input to a generative decoder architecture. This makes our framework a truly probabilistic generative approach for human motion prediction.

Recently, HP-GAN~\cite{BarsoumCVPRW2018} proposed a similar approach by utilizing advances in generative adversarial network (GAN) to model human motion prediction as a generative modeling task. But the authors have not evaluated its performance against the available deterministic state-of-the-art methods. The focus should be on the performance metric of long-term prediction to rule-out the phenomenon of convergence to mean pose sequence, which is evident in deterministic motion prediction methods~\cite{li2018convolutional}. However, the generative setup incorporated by HP-GAN does not have the flexibility for quality assessment of the generated motion. The challenge is to incorporate modifications in the probabilistic motion prediction model, which can offer a new direction to evaluate expressiveness of such frameworks for long-term prediction.

It has been shown that, the quality of predictions by a pure encoder-decoder setup is much better than a variational counterpart, mostly because of the complex objective - to generate novel samples (or to learn a continuous latent space) - of the latter. Hence, there has been an increasing interest  to incorporate a direct content loss (mean squared loss) on the available training samples even for generative modeling, as it ensures superior prediction quality alongside avoiding mode-collapse. Works like ~\cite{chen2016infogan,makhzani2015adversarial} incorporated autoencoder setup with generative adversarial objective to improve quality of generation with stabilized training regime. Motivated by this line of thought, unlike HP-GAN, the goal is to integrate direct content loss on the available full motion sequence (combined past and future frames) in the proposed conditional sequence generative framework. 
For each available full sequence, the proposed model should be able to predict the exact future sequence conditioned on the encoded past dynamics and some extrinsic latent representation. Moreover, as a given test sequence includes one of the plausible pose forecast dynamics, the latent random vector, along with modeling uncertainty in future prediction, must also be able to represent the exact pose forecast dynamics with utmost efficiency. 

Unlike HP-GAN, the proposed generative framework incorporates a novel conditional discriminator architecture. Here the discriminator not only acts like a critic, discriminating actual pose dynamics from the predicted ones; but also regresses the randomly sampled extrinsic vector, which was used for the prediction of the corresponding future dynamics. Design of such discriminator has two prominent traits. Firstly, it avoids the inherent problem of mode-collapse as it attempts to learn a one-to-one invertible mapping between the extrinsic latent vector and the corresponding motion prediction. Secondly, it offers a new way to enforce direct content loss (similar to deterministic encoder-decoder framework) on the prediction of probabilistic decoder output (more details in Approach Section). Thus, by integrating this novel modification to the discriminator architecture with an efficient learning algorithm (See Algorithm \ref{algo:1}), we are able to achieve superior motion prediction results as compared to previous methods. Such setup also provides a flexibility to compare quality of long-term prediction against previous deterministic state-of-the-art approaches.

\section{Related Works}
\label{sec:related-works}
Data-driven human motion prediction models have been explored by researchers for quite a along time in both computer animation and machine learning community. Before the deep-era various probabilistic graphical models have been tried to efficiently model human motion dynamics. Researchers have used time-series learning methods like Hidden Markov Model~\cite{arikan2003motion}, restricted Boltzmann machines~\cite{taylor2007modeling}, Gaussian process~\cite{wang2008gaussian}, switching linear dynamical system~\cite{pavlovic2001learning} to model human pose sequence data. However these methods fail to model the high-dimensional complex human pose sequence information effectively. Because of the highly nonlinear dependencies arose by the uncertainty in human movement, individually modeling various different factors affecting motion prediction does not scale well. These methods also suffer from complex training regime ~\cite{taylor2007modeling} with complicated inference pipeline as a result of the acquired sampling technique.

On the other hand, success of recurrent neural network (RNN) for modeling time-series data motivated researchers to effectively apply such architectures on human motion prediction task. Multitude of recent works ~\cite{fragkiadaki2015recurrent,Martinez_2017_CVPR} successfully used variants of recurrent sequence-to-sequence architecture to model complex human skeleton dynamics. Such methods consider a seed motion sequence of certain time-step to condition prediction of future pose dynamics by employing an encoder-decoder recurrent pipeline. Ghosh \etal~\cite{ghosh2017learning} employ an additional non-recurrent encoder and decoder to explicitly leverage spatial structure and dependencies between joint locations to improve prediction quality of human pose sequence. Jain \etal~\cite{jain2016structural} proposed Structural-RNN to exploit the underlying spatio-temporal graph for modeling human skeleton dynamics. However all these methods do not consider the stochasticity in future pose dynamics by modeling it as a deterministic prediction problem. Hence, expressiveness of these approaches in modeling long-term motion sequence deteriorates as a result of convergence to a mean pose sequence.

HP-GAN~\cite{BarsoumCVPRW2018} first attempted to model human motion prediction as a probabilistic generative approach. They leverage recent advances in generative adversarial network (GAN) ~\cite{goodfellow2014generative} to adversarially train a recurrent motion prediction framework. However, they fail to assess expressiveness of such generative approach against deterministic counterparts. In contrast, the proposed \textit{BiHMP-GAN} incorporates novel modifications in architecture and training regime to improve expressiveness of the probabilistic method against available deterministic approaches.

\begin{figure*}
\begin{center}
	\includegraphics[width=1.01\linewidth]{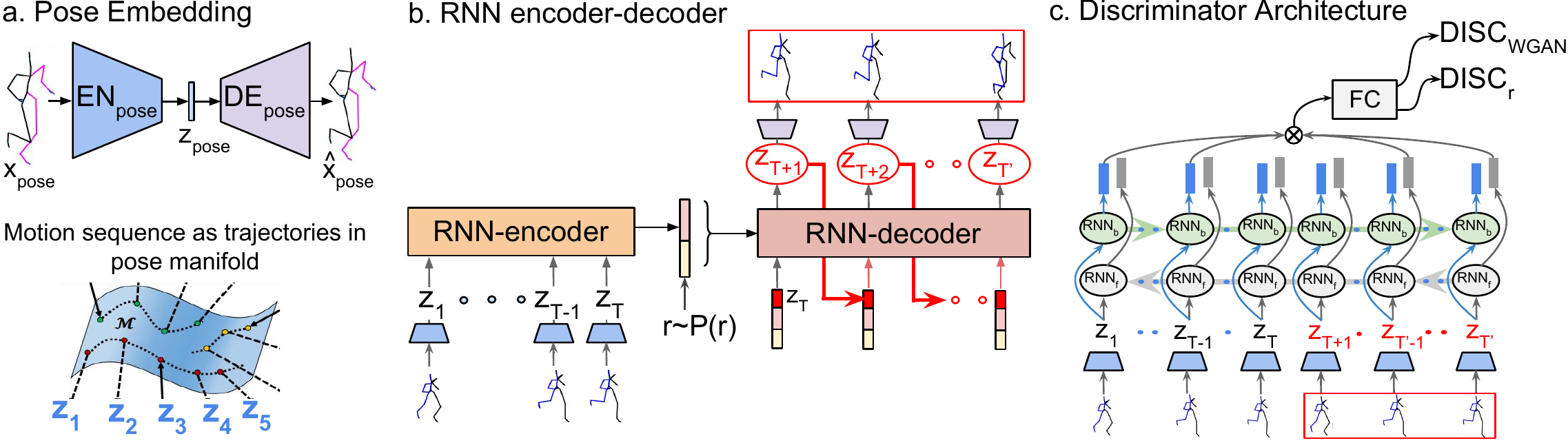}
	\caption{Illustration of the full \textit{BiHMP-GAN} pipeline. 
    Note that, $RNN_{dec}$ is modeled in a residual setup, where each cell $t$ predicts $\Delta\hat{z}_{t}$ which is added with $\hat{z}_{t-1}$ to obtain final prediction $\hat{z}_{t}$.}
 	\label{fig:fig_2}    
    \vspace{-4mm}
\end{center}
\end{figure*}

\section{Approach}
\label{sec:approach}
We here describe the details of the proposed probabilistic human motion prediction framework. 
The sequence prediction model takes a stream of input pose frames, which is considered as past motion conditioning. Let $X_{1:T} = [x_1, x_2,...x_T]$ be the sequence of input 3D pose representations for time-step $t=1$ to $T$. Here, a single pose frame is represented by a set of joint angle parameters in the kinematic representation form. Similarly, the output motion sequence is represented by $X_{T+1:T^\prime}$, where $(T^{\prime}-T)$ is the length of predicted sequence. The objective is to learn $P(X_{T+1:T^\prime}|X_{1:T})$, i.e. the model should predict future pose dynamics conditioned on a given past motion sequence.

The prime complexity in the generation of human motion sequence can be analyzed in two folds. Firstly, the generative model should predict plausible human pose representation at each time-step. Understanding the joint angle limits while generating a 3D human pose can be considered as the most important trait to avoid prediction of implausible joint angles. Secondly, the sequence of pose dynamics should be coherent to resemble like a real human motion dynamics. Previous methods do not address this complexities in human-motion modeling individually. A single recurrent network is employed for human motion prediction, as a black-box, to handle both the above complexities in the output motion prediction. Diverted from this general trend, we plan to first learn a continuous pose embedding space independent of the motion dynamics to avoid prediction of unlikely or improbable skeleton joint parameters. This is crucial, especially for models targeting long-term motion prediction, as short-term motion for less than 200ms constitutes minimal diversity in the forecasted pose with respect to the immediate past frames.

\subsection{Learning of Pose Embedding Representation}
The objective is to learn a pose embedding space, $z_{pose}$ so that $P(z_{pose})$ models the distribution of only plausible joint angle arrangements. The first choice is to use a generative adversarial network to model the same, which will include a pose generator (or decoder), $DE_{pose}$ as a transformation from $z_{pose}\sim P(z_{pose})$ to the actual skeletal pose, $x_{pose}\sim P(x_{pose})$. We emphasize learning of a generative model instead of a simple auto-encoder as the objective is to learn a continuous pose embedding space, which can allow effective interpolation of pose sequence between two plausible pose frames~\cite{radford2015unsupervised}. A simple autoencoder without explicit enforcement of being generative leads to learning of a discrete pose embedding space modeling only the available training samples, and hence delivers sub-optimal interpolation results. The core idea is to interpret pose sequence in later stage of the motion prediction framework, as a trajectory in the pose embedding space. Such setting not only enforces prediction of plausible pose frames, but also reduces burden on the subsequent sequence learning framework by segregating the complex task of efficient pose sequence prediction. 

Following the idea of modeling human motion as a trajectory in the pose embedding space, the pose sequence decoder network must output $z_{pose}$ sequence instead of $x_{pose}$ sequence directly, as attempted by previous approaches~\cite{li2018convolutional}. Similarly, the pose sequence encoder architecture would also take $z_{pose}$ sequence as input representation. This asks for an inference function to transform $x_{pose}$ to the corresponding $z_{pose}$, which is realized by introducing a pose encoder, $EN_{pose}$. Motivated from adversarial auto-encoder framework ~\cite{makhzani2015adversarial}, we train the full adversarial pose autoencoder by employing a pose discriminator, which can distinguish between predicted and actual skeletal joint angle patterns. Cyclic reconstruction loss is added on both $x_{pose}$ and $z_{pose}$ to enforce learning of an one-to-one mapping in a generative adversarial setup. 

\vspace{-4mm}
\begin{eqnarray*}
\mathcal{L}_{cyc} = \vert x_{pose} - \hat{x}_{pose} \vert + \vert z_{pose} - \hat{z}_{pose} \vert \\
Where,\: \hat{x}_{pose} = DE_{pose}(EN_{pose}(x_{pose})) \\and\: \hat{z}_{pose} = EN_{pose}(DE_{pose}(z_{pose}))
\end{eqnarray*}

Here, $z_{pose}$ is sampled from a predefined prior distribution $P(z_{pose})$.
Note that, $EN_{pose}$ is trained using only $\mathcal{L}_{cyc}$ loss, whereas $DE_{pose}$ is trained using $\mathcal{L}_{cyc} + \lambda\mathcal{L}_{adv}$.

Furthermore, effectiveness of the model is evaluated by visualizing interpolation results between two randomly chosen pose frames. A balance between the cyclic reconstruction loss, $\mathcal{L}_{cyc}$  and the adversarial discriminator loss, $\mathcal{L}_{adv}$ is maintained by exploring an effective relative weighting scheme. This is crucial, as more emphasize on cyclic reconstruction loss may derail the the setup towards learning a discrete embedding space with deteriorated generalization on novel pose samples.

\subsection{Probabilistic Motion Prediction Framework}
After obtaining an effective pose descriptor from the learned pose embedding space, we focus on modeling the temporal aspect of pose dynamics. Different human motion categories will form a certain type of trajectory in the learned pose embedding manifold. Note that, the trajectory should constitute smooth transitions of $z_{pose}$ as a result of the probabilistic generative approach to train the embedding representation. The resultant transformation functions viz. $EN_{pose}$ and $DE_{pose}$  with frozen learned parameters is utilized in later stage to effectively model human motion as a trajectory in the learned pose embedding. This is realized by introducing a recurrent sequence encoder  $RNN_{enc}$ and a decoder $RNN_{dec}$ architecture as shown in Figure \ref{fig:fig_2}.

$RNN_{enc}$ takes a sequence of pose embeddings as input, which can be represented as $Z_{1:T} = [z_1, z_2,...z_T] = [EN_{pose}(x_1), EN_{pose}(x_2),...EN_{pose}(x_T)]$. The final hidden sate representation at time $T$, i.e. $h^{enc}_T$ is considered as an intrinsic factor required for the prediction of future pose dynamics. To model the inherent stochasticity in the generation of future pose sequence, we introduce an extrinsic factor $r$. Here $r$ is considered as a random vector drawn from a probability distribution, $P(r)$, which can be taken as either Gaussian or Uniform prior distribution. To influence the prediction of future pose sequence the decoder recurrent network ($RNN_{dec}$) takes a tuple of both extrinsic and intrinsic factors, i.e. $(h^{enc}_T, r)$ as shown in Figure \ref{fig:fig_2}.

Previous approaches design the decoder as an autoregressive framework, which mostly considers short-term past sequence to regress the next pose representation. An optimum setup would be the one, where the next frame is directly influenced by both long-term and short-term past representations. Here, the long-term information is related to the global properties of the given past motion dynamics. This includes motion category and other pose and environmental constraints. Whereas, short-term representation constitutes pose dynamics from the immediate past pose enforcing smoothness in the predicted sequence. Motivated by this, we plan to feed a concatenated representation of $h^{enc}_T$, $r$ and the chained input from the predicted past pose, as input to the $RNN_{dec}$ at each time-step. Let the predicted sequence output from $RNN_{dec}$ be, $\hat{Z}_{T+1:T^\prime} = [\hat{z}_{T+1}, \hat{z}_{T+2},...\hat{z}_{T^\prime}]$. (Note that, here $RNN_{dec}$ is modeled in a residual setup, where each cell $t$ predicts $\Delta\hat{z}_{t}$ which is added with $\hat{z}_{t-1}$ to obtain final prediction $\hat{z}_{t}$). Then, the input at $t$th time-step to $RNN_{dec}$ will be a concatenated tuple of $(h^{enc}_T, r, \hat{z}_{t-1})$ as shown in Figure \ref{fig:fig_2}. The initial hidden state for $RNN_{dec}$ is also a function of both $h^{enc}_T$ and $r$. As the sequence decoder predicts the embeddings of actual pose representation, the final human pose prediction is obtained by utilizing the frozen $DE_{pose}$ transformation. Therefore the final output, $\hat{X}_{T+1:T^\prime} = [DE_{pose}({\hat{z}}_{T+1}), DE_{pose}(\hat{z}_{T+2}),...DE_{pose}(\hat{z}_{T^\prime})]$.

\subsubsection{Discriminator Design Supporting Enforcement of Content Loss} 
The sequence prediction framework is also designed by taking motivations from generative adversarial network. The objective is to enable modeling of variations in prediction of future sequence conditioned on the given past motion i.e. $P(X_{T+1:T^\prime}|X_{1:T})$. As described above, $RNN_{dec}$ effectively takes two input representations, viz. output of $RNN_{enc}$ and $r$. Following this, the discriminator takes the predicted pose sequence along with the input conditioned past frames as shown in Figure \ref{fig:fig_2}. Here, the discriminator architecture has 2 output heads, viz, a)$DISC_{WGAN}$ and b) $DISC_{r}$. The discriminator not only outputs  a single neuron for the usual adversarial loss, but also predicts the random $r$ vector which is being used to generate the corresponding predicted sequence.  

\begin{figure}[!tbp]
\begin{center}
	\includegraphics[width=1.0\linewidth]{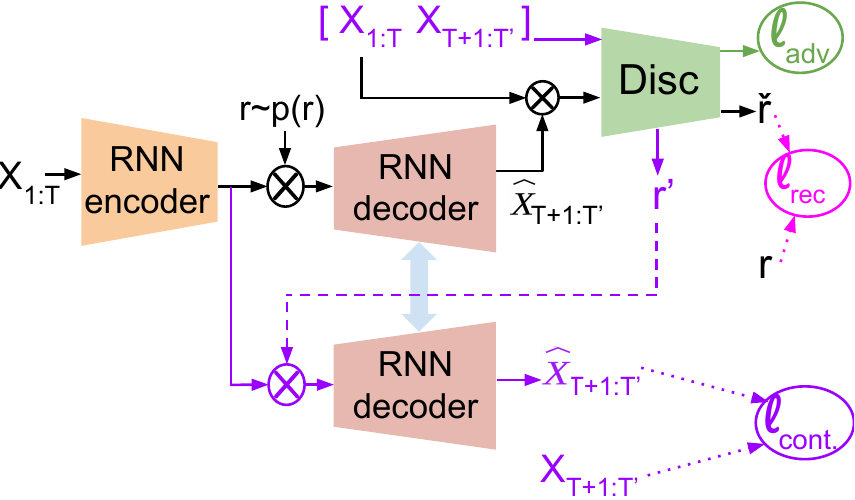}
	\caption{Workflow illustrating enforcement of content loss in \textit{BiHMP-GAN} as indicated by purple arrows.}
 	\label{fig:fig_3}    
\end{center}
\vspace{-4mm}
\end{figure}

Moreover, a separate critic network is introduced with similar architecture with a single output-head named as $DISC_{critic}$.
A binary cross entropy loss is applied on the output of $DISC_{critic}$ after the final sigmoid nonlinearity to learn a discriminative function to distinguish between the predicted and actual pose sequence. The single neuron output of $DISC_{WGAN}$ is used to enforce minimization of Earth Mover Distance (EMD) as proposed by Arjovsky \etal~\cite{arjovsky2017wasserstein}. Note that adversarial loss from only $DISC_{WGAN}$ is used to train the RNN encoder-decoder parameters for learning stability; following implementation tricks by Gulrajani \etal~\cite{gulrajani2017improved}. The additional output-head $DISC_{r}$ attached to the discriminator, is a novel approach to regress the $r$ vector, which can generate the input future sequence given the past motion dynamics. The prime motivation behind incorporation of $DISC_{r}$ can be of two folds. First, being able to regress $r$ while training the encoder-decoder parameters enforces learning of an one-to-one mapping avoiding mode-collapse. Secondly, it offers a new direction to enforce content information directly on the predicted motion sequence. Consider, there exists a particular $r$ which can generate the future frames exactly as it is given in a chosen training sample of length $T^\prime$. Now to be able to enforce a content loss directly on the predicted sequence of $RNN_{dec}$, we first perform an inference of the full sequence (of length $T^\prime$) through the trained discriminator to obtain a specific $r^\prime$ vector from the output head $DISC_{r}$. This $r^\prime$ is later utilized in the next iteration to enforce a direct content loss between predicted final pose sequence, $\hat{X}_{T+1:T^\prime}$ and the ground-truth, $X_{T+1:T^\prime}$ as described in Figure \ref{fig:fig_3}.

\begin{algorithm}[!b]
\label{algo:1}
\SetAlgoLined
        /*Initialization of parameters */\\
$\Theta_{ENC}$ : Parameters of $RNN_{enc}$\\
$\Theta_{DEC}$  : Parameters of $RNN_{dec}$\\
$\Theta_{DISC}$ : Parameters of $RNN_{disc}$\\
\vspace{1mm}
\For {$k$ iterations}{
 \For {$m$ steps}{
     $X_{1:T+\alpha T^\prime}$: minibatch training motion sequence\\
     $r$: random minibatch sampled from prior p(r)\\
     \vspace{1mm} $\hat{X}_{T+1:T^\prime}^r =  RNN_{dec}( \;RNN_{enc}( X_{1:T})\Vert r \;) $ \\
     \vspace{1mm} $\mathcal{L}_{adv}^{disc} = DISC_{WGAN}(\; X_{1:T}\Vert \hat{X}^r_{T+1:T^\prime} \;) - 
     DISC_{WGAN}(\; X_{1:T}\Vert {X}_{T+1:T^\prime} \;)$ \\
     \vspace{1mm} $ \mathcal{L}^{r}_{rec} = \vert r - DISC_r(\; X_{1:T}\Vert \hat{X}_{T+1:T^\prime}^r \;) \vert$ \\
     $   \; $ \\ /* Update parameters for $DISC$ network*/\\
    \vspace{1mm}
    $\Theta_{DISC} := \underset{\Theta_{DISC}}
    {\textrm{argmin}}\hspace{1mm}(  \mathcal{L}_{adv}^{disc} + \lambda_r \mathcal{L}^{r}_{rec})$\\
 }
 \vspace{1mm} $r^\prime = DISC_r(\; X_{1:T}\Vert X_{T+1:T^\prime} \;)$\\
 \vspace{1mm} $\hat{X}^{r^\prime}_{T+1:T^\prime} = RNN_{dec}( \;RNN_{enc}( X_{1:T})\Vert r^\prime \;)$\\
 \vspace{1mm} $\mathcal{L}^{X}_{content} = \vert X_{T+1:T^\prime} - \hat{X}^{r^\prime}_{T+1:T^\prime} \vert $\\ 
 \vspace{1mm} $\mathcal{L}_{adv}^{gen} = - DISC_{WGAN}(\; X_{1:T}\Vert \hat{X}^r_{T+1:T^\prime} \;)$ \\
  $   \; $ \\
 /* Update parameters of $RNN_{enc}$ and $RNN_{dec}$ */\\
 \vspace{1mm}
 $\Theta_{DEC} := \underset{\Theta_{DEC}}
 {\textrm{argmin}}\hspace{1mm}(  \mathcal{L}_{adv}^{gen} + \lambda_{r}\mathcal{L}^{r}_{rec} + 			\lambda_{c}\mathcal{L}^{X}_{content})$\\
  \vspace{1mm}
 $\Theta_{ENC} := \underset{\Theta_{ENC}}
  {\textrm{argmin}}\hspace{1mm}(  \mathcal{L}_{adv}^{gen} + \lambda_{c}\mathcal{L}^{X}_{content})$\\
 
 }
 
\caption{Training algorithm for \textit{BiHMP-GAN}, with explicit enforcement of direct content loss.}
\end{algorithm}


\subsubsection{Regularization by Recursive Prediction}
To further regularize the training procedure, we incorporate recursive prediction of motion sequence. Consider an input motion sequence, $X_{1:\alpha T^\prime}$ of length $\alpha T^\prime$, where $\alpha$ is some integer value depending on the available sequence length of a training sample. First, the prediction framework is used to obtain $\hat{X}_{T+1:T^\prime}$ by considering $X_{1:T}$ as past motion sequence. Following this, $\hat{X}_{T^\prime+1:2T^\prime}$ is obtained for $\alpha = 2$ by conditioning on the predicted past sequence, i.e. $\hat{X}_{T^\prime-T+1:T^\prime}$ as input dynamics. In general for a particular $\alpha$ value, $\hat{X}_{(\alpha-1)T^\prime+1:\alpha T^\prime}$ is obtained by considering the intrinsic input factor as a function of $\hat{X}_{(\alpha-1)T^\prime-T+1:(\alpha-1)T^\prime}$. But as discussed above, specific intrinsic factor $r_{\alpha}$ is required for each $\alpha$ value to be able to enforce a direct content loss in the probabilistic framework. $r_{\alpha}$ is obtained from the discriminator head, $DISC_{r}$ for the following concatenated input sequence: 
$[{X}_{(\alpha-1)T^\prime-T:(\alpha-1)T^\prime}, X_{(\alpha-1)T^\prime+1:\alpha T^\prime}]$ for each recursive $\alpha$ step. This regularization not only improves our long-term prediction results, but also acts like an effective solution to avoid convergence to mean pose unlike previous state-of-the-arts.

\subsubsection{Discriminator architecture}
HP-GAN proposes to utilize the full motion length $X_{1:T}\Vert X_{T+1:T^\prime}$ as input to the recurrent pose discriminator architecture ($\Vert$ represents concatenation operation). The goal is to match $P(X_{T+1:T^\prime}|X_{1:T})$ distribution with $P(\hat{X}^r_{T+1:T^\prime}|X_{1:T})$ for some $r\sim p(r)$ by following the generative adversarial learning technique. Unlike HP-GAN, to effectively capture $P(\hat{X}^r_{T+1:T^\prime}|X_{1:T})$ we propose certain intuitive modifications  to the discriminator architecture. The qualitative results of HP-GAN~\cite{BarsoumCVPRW2018} clearly highlights the catastrophic drift in the initial pose predictions as compared to the immediate past. In general, by effectively modeling $P(X_{T+\tau} | X_{1:T})$ for a very small $\tau$ (i.e. less than 50ms) such spurious drifts can be avoided in the predicted motion sequence.
This way, we enforce the model to learn less diversity in the prediction of initial $\tau$ frames for any $r\sim p(r)$, extrinsic factor, hence avoiding the catastrophic drift in the generations. Following this we model both   $P(X_{T+\tau} | X_{1:T})$ and $P( X_{T-1-\tau}| X_{T+1:T^\prime})$ by employing a bidirectional recurrent neural network as shown in Figure \ref{fig:fig_2}. We utilize the idea of plausible trajectory in the learned pose embedding, by feeding the sequence of pose embedding representations (i.e. the output of $EN_{pose}$) to the bidirectional recurrent architecture. Final output of the discriminator is extracted from 4 different hidden representations i.e. final hidden state of both forward and backwards recurrent RNN along with $h_{forwrd}(T+\tau)$ and $h_{backward}(T-1-\tau)$ as shown in Figure \ref{fig:fig_2}.


\begin{table*}
  \centering
  \caption{\small Comparison of motion prediction error on Human 3.6M dataset for short-term (80ms, 160ms, 320ms, 400ms) and long-term(1000ms) prediction.
  \textit{BiHMP-GAN} clearly outperforms others in long-term prediction.}
  \label{table1}
  \small
  \setlength{\tabcolsep}{3pt}
  \begin{tabularx}{0.99\textwidth}{cccccc|ccccc|ccccc|ccccc}
    \noalign{\hrule height 1pt}      
    &\multicolumn{5}{c}{Walking}\vline
    &\multicolumn{5}{c}{Eating}\vline
    &\multicolumn{5}{c}{Smoking}\vline
    &\multicolumn{5}{c}{Discussion}\\
    ms & 80 & 160 & 320 & 400 & 1000  
       & 80 & 160 & 320 & 400 & 1000  
       & 80 & 160 & 320 & 400 & 1000  
       & 80 & 160 & 320 & 400 & 1000\\
    \noalign{\hrule height 0.5pt}
    \scriptsize   RRNN 
      & 0.33 & 0.56 & 0.78 & 0.85 & 1.14  
      & 0.26 & 0.43 & 0.66 & 0.81 & 1.34  
   	  & 0.35 & 0.64 & 1.03 & 1.15 & 1.83 
      & 0.37 & 0.77 & 1.06 & 1.10 & {1.79}\\
    \scriptsize   Conv-Motion 
    	& \textbf{0.33} &  0.54 &  0.68 & {0.73} & {0.92}  
        & {0.22} & {0.36} & {0.58} & {0.71} & {1.24}  
        & \textbf{0.26} & {0.49} &  0.96 & {0.92} & {1.62} 
        & \textbf{0.32} &  0.67 &  0.94 & {1.01} & 1.86\\
    \noalign{\hrule height 0.5pt}        
    \scriptsize  \textit{HP-GAN($min_{err}$)}
    	& {0.95} &  1.17 &  1.69 & 1.79 & 2.47  
        & {1.28} & {1.47} & {1.70} & {1.82} & {2.51}  
        & {1.71} & {1.89} &  2.33 & {2.42} & {3.2} 
        & {2.29} &  2.61 &  2.79 & 2.88 & 3.67\\
    \scriptsize   $Ours(min_{err})$ 
    	& \textbf{0.33} &  \textbf{0.52} &  0.64 & {0.69} & {0.88}  
        & {0.21} & \textbf{0.33} & {0.55} & {0.71} & \textbf{1.20}  
        & {0.26} & \textbf{0.49} &  \textbf{0.91} & {0.88} & {1.12} 
        & \textbf{0.32} &  0.65 &  0.92 & {9.98} & 1.78\\
     \scriptsize   $Ours(r^\prime)$ 
    	& \textbf{0.33} & \textbf{ 0.52 }&  \textbf{0.63} & \textbf{0.67} & \textbf{0.85} 
        & \textbf{0.20} & \textbf{0.33} & \textbf{0.54} & \textbf{0.70} & \textbf{1.20}  
        & \textbf{0.26} & {0.50} &  \textbf{0.91} & \textbf{0.86} & \textbf{1.11} 
        & {0.33} &  \textbf{0.65} &  \textbf{0.91} & \textbf{9.95} & \textbf{1.77}\\
     
    \hline
  \end{tabularx}
  \vspace{-2mm}
\end{table*}
\begin{table}[t]
	\centering
	\caption{\small Ablation analysis on Human 3.6M, reporting mean average error (across 15 categories) at 1000ms}
	\begin{tabular}{ccc}
		\hline
		Metrics & $r^\prime=DISC_r$ & $r^\prime=argmin_{err}$  \\  
		\hline      
		\scriptsize{Without pose embedding} & {1.76} & {1.76} \\ 
		\scriptsize{Without encoder state in chaining} & 1.71 & 1.72 \\ 
        \scriptsize{Without recursive prediction} & 1.69 & 1.69 \\ 
        \scriptsize{\textit{BiHMP-GAN}} & \textbf{1.67} & \textbf{1.68} \\ 
		\hline
	\end{tabular}\label{table2}
    \vspace{-2mm}
\end{table}

\begin{table}[t]
	\centering
	\caption{\small Quantitative comparison with HP-GAN (classifier accuracy on real test samples of Human 3.6M: 55.4\%). We use the proposed discriminator architecture to design critic network for HP-GAN, which can easily detect the catastrophic drift in the initial predicted sequence}
	\begin{tabular}{ccc}
		\hline
		Accuracy  & Motion Classifier & Critic  \\  
		\hline  
        \scriptsize{HP-GAN} & {9.8} & {18.5} \\ 
        \scriptsize{\textit{BiHMP-GAN}} & \textbf{41.2} & \textbf{74.6} \\ 
		\hline
	\end{tabular}\label{table3}
    \vspace{-2mm}
\end{table}

\section{Experiments}
In this section we describe experimental details of \textit{BiHMP-GAN} along with analysis of both qualitative and quantitative results on two publicly available datasets; viz. a) Human 3.6M~\cite{ionescu2014human3} and CMU MOCAP. 

The full pipeline of \textit{BiHMP-GAN} is implemented in tensorflow with ADAM optimizer. We use a batch size of 32 with learning rate set at 0.00005. Single layer LSTM~\cite{chung2014empirical} with 512 hidden units is incorporated as a recurrent architecture for both sequence encoder, decoder and bidirectional discriminator network. Following previous motion prediction works ~\cite{li2018convolutional,Martinez_2017_CVPR} the length of intrinsic past pose sequence is set to 50, i.e. 2 seconds of skeleton motion at 25 fps setting. Considering fair evaluation on long-term prediction, the length of predicted motion sequence is set to 25. We choose $\tau$=1 for the modified discriminator architecture. The value of $\alpha$ for the recursive prediction regularization is set to 2. 
Instead of training the recurrent encoder-decoder parameters with addition of all the loss functions described above, we sequentially iterate over $\mathcal{L}_{content}^X$ and the recursive content regularization loss separately from the adversarial loss, $\mathcal{L}_{adv}^{disc} + \lambda_r \mathcal{L}^{r}_{rec}$ by defining different ADAM optimizers for each of them. We choose $\mathcal{N}(0, 1)$ prior distribution for both $z_{pose}$ and $r$ with $32$ and $8$ dimensions respectively. To ensure fair comparison, we trained \textit{HP-GAN}~\cite{BarsoumCVPRW2018} on Human 3.6M dataset with the same setting of sequence lengths and input representations using the publicly available implementation. 

\subsection{Datasets}
Human 3.6M is a widely accepted dataset for benchmarking human motion prediction works as it constitutes highly diverse action categories with actions performed by multiple subjects. 
Preprocessing and data selection criteria is directly followed from the recent work of Li~\etal~\cite{li2018convolutional}. We finally use a 54 dimensional input representation as $x_{pose}$ eliminating global orientation and translation parameters. Euclidean error on the predicted Euler angles is considered as an evaluation metric for comparison of \textit{BiHMP-GAN} against previous state-of-the-art motion prediction methods.

We also report performance of \textit{BiHMP-GAN} on CMU motion capture dataset to demonstrate generalization of the proposed probabilistic prediction method. We follow the preprocessing and data selection criteria from  Li~\etal~\cite{li2018convolutional}, which finally selects eight action categories after pruning interaction based and other repeated action categories. 

\begin{table*}[h!]
  \centering
  \caption{\small Comparison of motion prediction error on CMU MOCAP dataset for short-term (80ms, 160ms, 320ms, 400ms) and long-term(1000ms) prediction. \textit{BiHMP-GAN} clearly outperforms others in long-term prediction.} 
  \label{table4}
    \small
  \setlength{\tabcolsep}{3pt}
  \begin{tabularx}{0.99\textwidth}{cccccc|ccccc|ccccc|ccccc}
     \noalign{\hrule height 1pt}  
    &\multicolumn{5}{c}{Basketball}\vline 
    &\multicolumn{5}{c}{Basketball Signal}\vline
    &\multicolumn{5}{c}{Directing Traffic}\vline
    &\multicolumn{5}{c}{Jumping}\\
    ms & 80 & 160 & 320 & 400 & 1000  
       & 80 & 160 & 320 & 400 & 1000  
       & 80 & 160 & 320 & 400 & 1000  
       & 80 & 160 & 320 & 400 & 1000\\\hline
    \scriptsize{RRNN}
    	& 0.50 & 0.80 & 1.27 & 1.45 & \bf1.78  
        & 0.41 & 0.76 & 1.32 & 1.54 & 2.15  
        & 0.33 & 0.59 & 0.93 & 1.10 & 2.05  
        & 0.56 & 0.88 & 1.77 & 2.02 & 2.40\\
    \scriptsize{Conv-Motion} 
    	& 0.37 & 0.62 & 1.07 & 1.18 & 1.95  
        & \textbf{0.32} & 0.59 & 1.04 & 1.24 & 1.96         
        & \textbf{0.25} & 0.56 & 0.89 & 1.00 & 2.04  
        & 0.39 & 0.60 & 1.36 & 1.56 & 2.01\\\hline
        
    \scriptsize{Ours($min_{err}$)} 
    	& \textbf{0.36} & \textbf{0.60} & 1.02 & 1.12 & 1.84  
        & 0.33 & \textbf{0.56} & \textbf{1.00} & 1.19 & 1.89         
        & \textbf{0.25} & 0.52 & \textbf{0.84} & \textbf{0.96} & 1.97
        & \textbf{0.38} & \textbf{0.57} & 1.32 & 1.51 & 1.94   \\
     \scriptsize{Ours($r^\prime$)} 
    	& 0.37 & 0.62 & \textbf{1.01} & \textbf{1.11} & 1.83  
        & \textbf{0.32} &\textbf{ 0.56} & 1.01 &\textbf{ 1.18} & \textbf{1.88}         
        & \textbf{0.25} & \textbf{0.51} & 0.85 & \textbf{0.96 }& \textbf{1.95}  
        & 0.39 & \textbf{0.57} & \textbf{1.31} & \textbf{1.50} & \textbf{1.93}    \\

    \noalign{\hrule height 0.75pt}   
    &\multicolumn{5}{c}{Running}\vline
    &\multicolumn{5}{c}{Soccer}\vline
    &\multicolumn{5}{c}{Walking}\vline
    &\multicolumn{5}{c}{Washwindow}\\
    ms & 80 & 160 & 320 & 400 & 1000  
       & 80 & 160 & 320 & 400 & 1000  
       & 80 & 160 & 320 & 400 & 1000  
       & 80 & 160 & 320 & 400 & 1000\\\hline

    \scriptsize{RRNN}
    	& 0.33 & 0.50 & 0.66 & 0.75 & 1.00  
        & 0.29 & 0.51 & 0.88 & 0.99 & 1.72  
        & 0.35 & 0.47 & 0.60 & 0.65 & 0.88  
        & \textbf{0.30} & 0.46 &\textbf{ 0.72} & \bf 0.91 & 1.36\\
    \scriptsize{Conv-Motion} 
    	& 0.28 & 0.41 & 0.52 & 0.57 & 0.67  
        & \textbf{0.26} & 0.44 & 0.75 & 0.87 & 1.56  
        & 0.35 & \textbf{0.44} & 0.45 & 0.50 & 0.78  
        & \textbf{0.30} & 0.47 & 0.80 & 1.01 & 1.39\\\hline
 
    \scriptsize{Ours($min_{err}$)} 
    	& \textbf{0.27} & \textbf{0.40} & \textbf{0.49} & 0.54 & 0.65  
        & \textbf{0.26} &\textbf{ 0.43} &\textbf{ 0.71} & 0.84 & 1.52  
        & \textbf{0.34} & \textbf{0.44} & \textbf{0.43} & 0.47 &\textbf{ 0.71}  
        & \textbf{0.30} & 0.48 & 0.76 & 0.98 & 1.32\\  
     \scriptsize{Ours($r^\prime$)} 
    	& 0.28 & \textbf{0.40} & 0.50 & \textbf{0.53} & \textbf{0.62}  
        & \textbf{0.26} & 0.44 & 0.72 & \textbf{0.82} & \textbf{1.51  }
        & 0.35 & 0.45 & 0.44 & \textbf{0.46} & 0.72  
        & 0.31 & \textbf{0.46} & 0.77 & 0.92 & \textbf{1.31}\\   
	\hline
  \end{tabularx}
\end{table*}

\begin{figure*}[htp]
\begin{center}
	\includegraphics[width=1.01\linewidth]{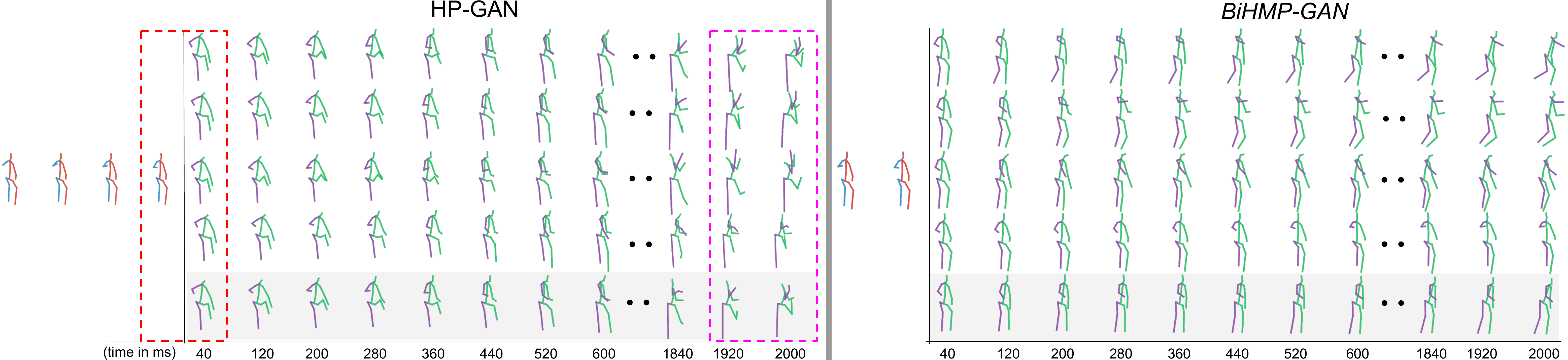}
	\caption{
    Qualitative results on Human 3.6M dataset on \textit{eating} category. It illustrates variations in forcasted motion (green-purple) for a given seed sequence (red-blue) as modeled by HP-GAN and \textit{BiHMP-GAN}. The last row shows the motion sequence generated via $min_{err}$ strategy. We highlight the catastrophic drift in the predicted motion of HP-GAN by dotted red box. We observe generation of unrealistic pose for long-term predictions by HP-GAN (highlighted in pink box), as it does not enforce generation of plausible pose frame. Also, generations of HP-GAN for a given seed sequence, lack variation for different latent vector $r$, as opposed to \textit{BiHMP-GAN}.}
    \vspace{-2mm}
    \label{fig:fig_4}  
\end{center}
\end{figure*}

\subsection{Comparison with other generative approaches}
We first compare our prediction performance against the available generative model \textit{HP-GAN}~\cite{BarsoumCVPRW2018}. After training \textit{HP-GAN} on the same settings for Human 3.6M dataset, efficacy of the predicted motion is evaluated by quantifying discriminability of a critic network to classify between the generated and real motion dynamics. Note that, we have employed the proposed modified discriminator architecture for the critic network to specifically consider the initial drift in predicted motion (see Table \ref{table3}). We also report performance of the generated motion by feeding the concatenated seed sequence and the generated motion to an action classifier trained only on real human motion dynamics(see Table \ref{table3}). Both qualitative (see Figure \ref{fig:fig_4}) and quantitative (see Table \ref{table3}) results clearly demonstrate superiority of \textit{BiHMP-GAN}. As a generative model, unlike \textit{HP-GAN}, \textit{BiHMP-GAN} is able to predict diverse prediction sequences without loosing the coherence with immediate past conditioning.

\subsection{Comparison with other deterministic approaches}
For each test sample $X_{1:T^\prime}$ of length $T^\prime$, there exist a particular $r^\prime$ which can model the exact predicted motion as $\hat{X}^{r^\prime}_{T+1:T^\prime} = RNN_{dec}(RNN_{enc}( X_{1:T})\Vert r^\prime)$. Therefore, modeling expressibility of a generative method can be evaluated by obtaining the best possible value of $r^\prime$ which can express a given test sample. Motivated by this, we define two different metrics to quantitatively assess the quality of non-deterministic predictions.

Firstly, considering $r^\prime= DISC_r( X_{1:T}\Vert X_{T+1:T^\prime})$, we report the prediction error of $\hat{X}^{r^\prime}_{T+1:T^\prime}$ against the corresponding ground-truth ${X}_{T+1:T^\prime}$, which is denoted as $Ours(r^\prime)$ in Table \ref{table1} and \ref{table4}. The metrics clearly demonstrate quality of the generated motion for both short-term (80 ms, 160 ms, 320 ms and 400 ms) and long-term prediction (1000 ms). Improved results on long-term prediction performance shows effectiveness \textit{BiHMP-GAN} in overcoming the phenomenon of convergence to mean pose, which is quite evident in deterministic approaches; RRNN~\cite{Martinez_2017_CVPR} and Conv-Motion~\cite{li2018convolutional}.

However, in the previous metric comparison, we have to use the ground-truth prediction $X_{T+1:T^\prime}$ as an input to the discriminator to obtain a particular vector $r^\prime$. Hence, we also propose another metric, to assess expressibility of the probabilistic motion prediction model as follows. We first save a batch of 1000 vectors $r_i$ randomly sampled from the prior distribution $P(r)$. Then, for each test sample $X_{1:T^\prime}$ we report the minimum Euclidean error as, $\min Error (\hat{X}^{r_i}_{T+1:T^\prime}, X_{T+1:T^\prime})$ for $i$=1,2,...1000. Table \ref{table1} and \ref{table4} holds comparison of this metric under the row heading $Ours(min_{err})$ and \textit{HP-GAN}($min_{err}$). It clearly highlights expressiveness of \textit{BiHMP-GAN} against \textit{HP-GAN} and other the deterministic approaches.

\subsection{Ablation study}
Here, we quantitatively analyze effectiveness of various design and learning schemes proposed for \textit{BiHMP-GAN}. To demonstrate the advantage of learning pose embedding representation, we compare \textit{BiHMP-GAN} against a baseline without the pose embedding transformations (see Table \ref{table2}). For the decoder setup, the effect of feeding concatenated previous pose feature along with the intrinsic encoder hidden state is evaluated against a baseline; with input sequence of only chained previous pose feature (See Table \ref{table2}). Finally, the effect of incorporating recursive prediction regularization in the training of \textit{BiHMP-GAN} is demonstrated against a baseline designed without any such regularization.

\section{Conclusion}
In this work, we proposed a novel probabilistic generative model for prediction of uncertain future motion dynamics. Being generative we have carefully designed the framework to model the available training sequences with a direct content loss. Modeling human motion as a trajectory in pose embedding makes \textit{BiHMP-GAN} devoid of generating unrealistic pose frames as compared to other approaches. We demonstrate improved expressibility of \textit{BiHMP-GAN} specially for long-term motion prediction against other deterministic motion prediction works. In future, we plan to extend similar training framework for complex motion sequences like, dance, martial arts etc. by aiming towards achieving a general motion embedding.

\section{Acknowledgements} This work was supported by a CSIR Fellowship (Jogendra), and a project grant from Robert Bosch Centre for Cyber-Physical Systems, IISc.

{\small
  \bibliographystyle{aaai}
  \bibliography{ms}
}

\end{document}